\documentclass[11pt,a4paper]{article}
\usepackage[hyperref]{acl2020}
\usepackage{pslatex}
\usepackage{latexsym}

\usepackage{url}
\usepackage{latexsym}
\usepackage{dirtytalk}
\usepackage{amsmath}
\usepackage{booktabs}
\usepackage{graphicx}
\usepackage{lipsum, color}
\usepackage{multirow}
\usepackage{array}
\usepackage{dirtytalk}
\usepackage{multicol}

\makeatletter
\g@addto@macro{\UrlBreaks}{\UrlOrds}
\makeatother
\usepackage{array}
\usepackage{graphbox}
\newcolumntype{L}[1]{>{\raggedright\let\newline\\\arraybackslash\hspace{0pt}}m{#1}}
\newcolumntype{C}[1]{>{\centering\let\newline\\\arraybackslash\hspace{0pt}}m{#1}}
\newcolumntype{R}[1]{>{\raggedleft\let\newline\\\arraybackslash\hspace{0pt}}m{#1}}
\usepackage{pifont}
\usepackage{amsmath,amsfonts,amssymb,amsthm}
\usepackage{mathtools}
\usepackage{commath}
\definecolor{tp1}{HTML}{66CC00}
\definecolor{tp2}{HTML}{009900}
\definecolor{tp3}{HTML}{999900}
\definecolor{tp4}{HTML}{FF0000}
\definecolor{tp5}{HTML}{FF6666}
\usepackage{subcaption}
\usepackage{caption}
\usepackage[inline]{enumitem}
\usepackage{caption, subcaption}
\usepackage{xargs}                      
\usepackage{tabularx}
\usepackage{graphicx,wrapfig,lipsum}
\usepackage{multirow}
\usepackage{amsmath}
\usepackage{color,soul}
\usepackage[colorinlistoftodos,prependcaption,textsize=scriptsize]{todonotes}
\usepackage{arydshln}
\usepackage{multicol}
\usepackage{soul}
\setuldepth{Summer}

\usepackage[belowskip=-5pt,aboveskip=5pt]{caption}
\usepackage{xspace}

\newcommand{\summer}{\textsc{Summer}\xspace}
\newcommand{\textrank}{\textsc{Text\-Rank}\xspace}
\newcommand{\summarunner}{\textsc{Summa\-RuNNer}\xspace}
\newcommand{\summarunnerdif}{\textsc{Summa\-RuNNer*}\xspace}
\newcommand{\scenesum}{\textsc{Scene\-Sum}\xspace}

\aclfinalcopy 

\title{Screenplay Summarization Using Latent Narrative Structure}

\author{Pinelopi Papalampidi\textsuperscript{1} \quad
  Frank Keller\textsuperscript{1} \quad
  Lea Frermann\textsuperscript{2} \quad
  Mirella Lapata\textsuperscript{1} \\
  \textsuperscript{1}Institute for Language, Cognition and Computation \\
    School of Informatics, University of Edinburgh \\
   \textsuperscript{2}School of Computing and Information Systems \\
    University of Melbourne \\
  \url{p.papalampidi@sms.ed.ac.uk},~~\url{keller@inf.ed.ac.uk}, \\ 
  \url{lea.frermann@unimelb.edu.au},~~\url{mlap@inf.ed.ac.uk}
}

\date{}

\begin{document}
\maketitle

\renewcommand{\UrlFont}{\ttfamily\small}

\begin{abstract}
  Most general-purpose extractive summarization models are trained on
  news articles, which are short and present all important information
  upfront. As a result, such models are biased by position and often
  perform a smart selection of sentences from the beginning of the
  document. When summarizing long narratives, which have complex
  structure and present information piecemeal, simple position
  heuristics are not sufficient. In this paper, we propose to
  explicitly incorporate the underlying structure of narratives into
  general unsupervised and supervised extractive summarization
  models. We formalize \emph{narrative structure} in terms of key
  narrative events (turning points) and treat it as latent in order to
  summarize screenplays (i.e.,~extract an optimal sequence of
  scenes). Experimental results on the CSI corpus of TV screenplays,
  which we augment with scene-level summarization labels, show that
  latent turning points correlate with important aspects of a CSI
  episode and improve summarization performance over general
  extractive algorithms, leading to more complete and diverse
  summaries.
\end{abstract}

\section{Introduction}
\label{sec:introduction}

Automatic summarization has enjoyed renewed interest in recent years
thanks to the popularity of modern neural network-based approaches
\cite{cheng2016neural,nallapati2016abstractive,nallapati2017summarunner,zheng2019sentence}
and the availability of large-scale datasets containing hundreds of
thousands of document--summary pairs
\cite{nytcorpus,hermann-nips15,newsroom-naacl18,narayan-etal-2018-dont,fabbri-etal-2019-multi,liu-lapata-2019-hierarchical}. Most
efforts to date have concentrated on the summarization of news
articles which tend to be relatively short and formulaic following an
``inverted pyramid'' structure which places the most essential, novel
and interesting elements of a story in the beginning and supporting
material and secondary details afterwards.  The rigid structure of
news articles is expedient since important passages can be identified
in predictable locations (e.g.,~by performing a ``smart selection'' of
sentences from the beginning of the document) and the structure itself
can be explicitly taken into account in model design (e.g.,~by
encoding the relative and absolute position of each sentence).

\begin{figure}[t]
    \centering
    \includegraphics[width=\columnwidth,page=1]{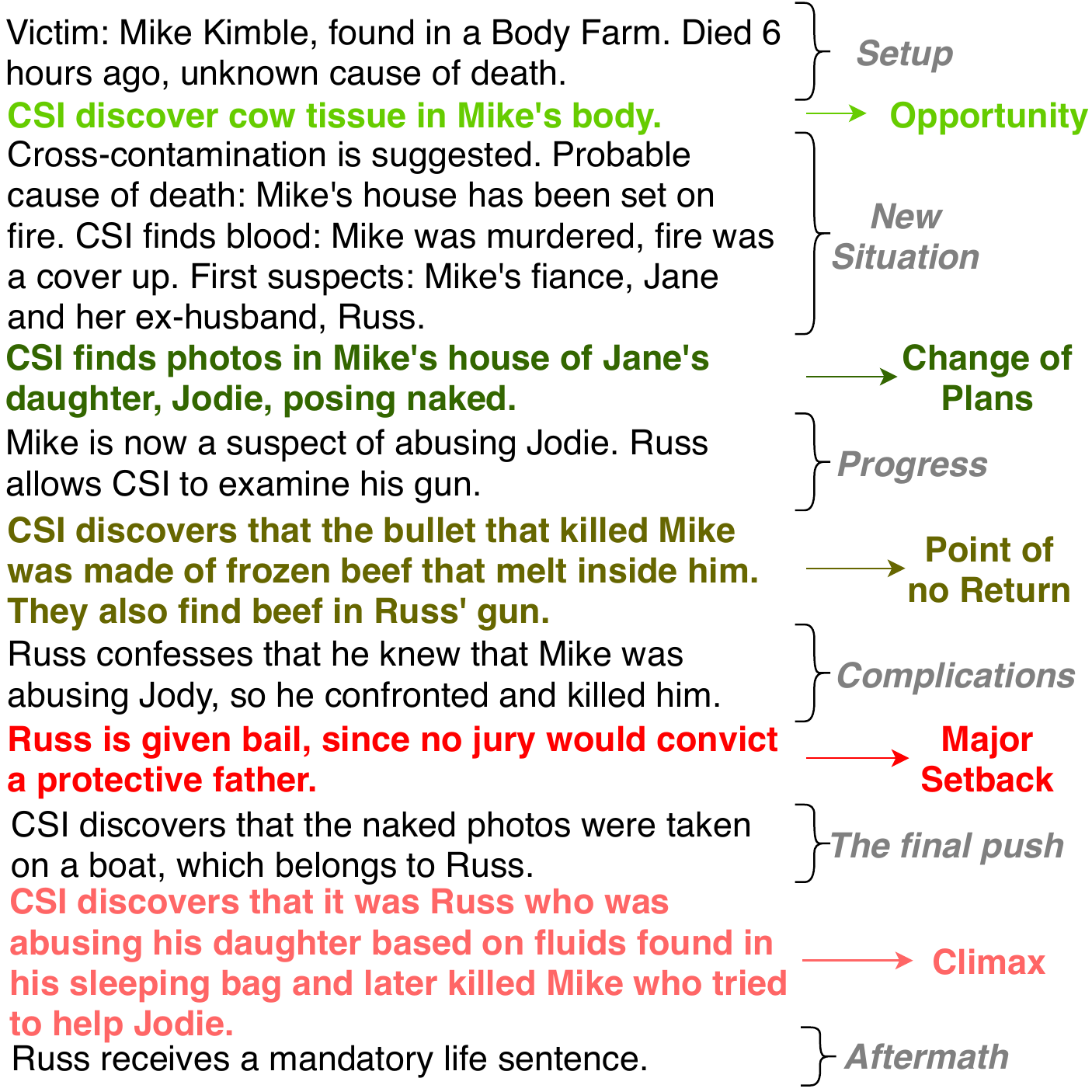}
    \caption{Example of narrative structure for episode ``Burden of
      Proof'' from TV series Crime Scene Investigation (CSI); turning
      points are highlighted in color.}
    \label{fig:tps_example}
\end{figure}

\begin{figure}[t]
    \centering
    \includegraphics[width=0.45\textwidth,page=1]{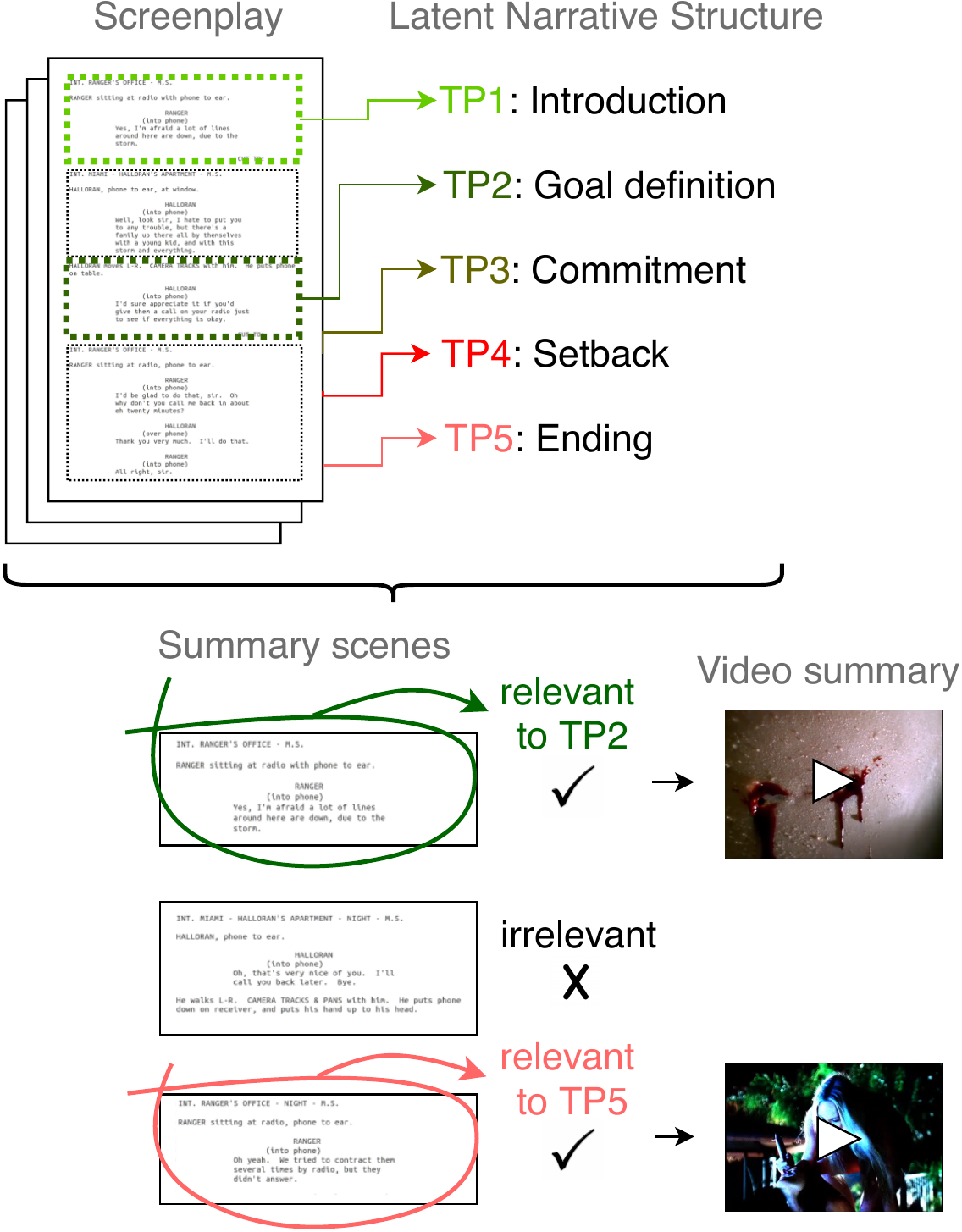}
    \caption{We first identify scenes that act as turning points
      (i.e.,~key events that segment the story into sections). We next
      create a summary by selecting informative scenes,
      i.e.,semantically related to turning points.}
    \label{fig:system-overview}
\end{figure}

In this paper we are interested in summarizing longer narratives,
i.e.,~screenplays, whose form and structure is far removed from
newspaper articles. Screenplays are typically between 110 and 120
pages long (20k words), their content is broken down into scenes, which contain
mostly dialogue (lines the actors speak) as well as descriptions
explaining what the camera sees. Moreover, screenplays are
characterized by an underlying \emph{narrative structure,} a sequence
of events by which a story is defined \cite{cutting2016narrative}, and
by the story's characters and their roles \cite{Propp:1968}. Contrary
to news articles, the gist of the story in a screenplay is not
disclosed at the start, information is often revealed piecemeal;
characters evolve and their actions might seem more or less important
over the course of the narrative. From a modeling perspective,
obtaining training data is particularly problematic: even if one could
assemble screenplays and corresponding summaries (e.g.,~by mining IMDb
or Wikipedia), the size of such a corpus would be at best in the range
of a few hundred examples not hundreds of thousands. Also note that
genre differences might render transfer learning \cite{Pan:ea:2010}
difficult, e.g.,~a model trained on movie screenplays might not
generalize to sitcoms or soap operas.

Given the above challenges, we introduce a number of assumptions to
make the task feasible. Firstly, our goal is to produce
\emph{informative} summaries, which serve as a surrogate to reading
the full script or watching the entire film. Secondly, we follow
\citet{gorinski-lapata-2015-movie} in conceptualizing screenplay
summarization as the task of identifying a sequence of informative
scenes. Thirdly, we focus on summarizing television programs such as
\textit{CSI: Crime Scene Investigation} \cite{frermann2018whodunnit}
which revolves around a team of forensic investigators solving
criminal cases. Such programs have a complex but well-defined
structure: they open with a crime, the crime scene is examined,
the victim is identified, suspects are introduced, forensic clues are
gathered, suspects are investigated, and finally the case is solved.

In this work, we adapt general-purpose extractive summarization
algorithms \cite{nallapati2017summarunner,zheng2019sentence} to
identify informative scenes in screenplays and  instill in
them  knowledge about narrative film structure
\cite{Hauge:2017,cutting2016narrative,freytag1896freytag}.
Specifically, we adopt a scheme commonly used by screenwriters as a
practical guide for producing successful screenplays. According to
this scheme, well-structured stories consist of six basic stages which
are defined by five \emph{turning points} (TPs), i.e.,~events which
change the direction of the narrative, and determine the story's
progression and basic thematic units. In Figure~\ref{fig:tps_example},
TPs are highlighted for a CSI episode. Although the link between
turning points and summarization has not been previously made, earlier
work has emphasized the importance of narrative structure for
summarizing books \cite{mihalcea2007explorations} and social media
content \cite{kimstoria}. More recently, \citet{papalampidi2019movie}
have shown how to identify turning points in feature-length
screenplays by projecting synopsis-level annotations.

Crucially, our method does not involve manually annotating turning
points in CSI episodes. Instead, we approximate narrative structure
automatically by pretraining on the annotations of the TRIPOD dataset
of \newcite{papalampidi2019movie} and employing a variant of their
model. We find that narrative structure representations learned on
their dataset (which was created for feature-length films), transfer
well across cinematic genres and computational tasks. We propose a
framework for end-to-end training in which narrative structure is
treated as a latent variable for summarization.  We extend the CSI
dataset \cite{frermann2018whodunnit} with binary labels indicating
whether a scene should be included in the summary and present
experiments with both supervised and unsupervised summarization
models. An overview of our approach is shown in
Figure~\ref{fig:system-overview}.

Our contributions can be summarized as follows: (a)~we develop methods
for instilling knowledge about narrative structure into generic
supervised and unsupervised summarization algorithms; (b)~we provide a
new layer of annotations for the CSI corpus, which can be used for
research in long-form summarization; and (c)~we demonstrate that
narrative structure can facilitate screenplay summarization; our
analysis shows that key events identified in the latent space
correlate with important summary content.

\section{Related Work}
\label{sec:related-work}

A large body of previous work has focused on the computational
analysis of narratives \cite{Mani:2012,Richards:ea:2009}. Attempts to
analyze how stories are written have been based on sequences of events
\cite{Schank:Abelson:1975,chambers-jurafsky-2009-unsupervised}, plot
units
\cite{mcintyre-lapata-2010-plot,goyal-etal-2010-automatically,Finlayson:2012}
and their structure \cite{Lehnert:1981,Rumelhart:1980}, as well as on
characters or personas in a narrative
\cite{black1979evaluation,Propp:1968,bamman-etal-2014-bayesian,bamman-etal-2013-learning,Vargas:ea:2014}
and their relationships
\cite{Elson:ea:2010,agarwal-EtAl:2014:CLFL,Srivastava:ea:2016}. 

As mentioned earlier, work on summarization of narratives has had
limited appeal, possibly due to the lack of annotated data for
modeling and evaluation.  \newcite{kazantseva2010summarizing}
summarize short stories based on importance criteria (e.g.,~whether a
segment contains protagonist or location information); they create
summaries to help readers decide whether they are
interested in reading the whole story, without revealing its
plot. \newcite{mihalcea2007explorations} summarize books with an
unsupervised graph-based approach operating over segments
(i.e.,~topical units).  Their algorithm first generates a summary for
each segment and then an overall summary by collecting
sentences from the individual segment summaries.

Focusing on screenplays,
\newcite{gorinski-lapata-2015-movie} generate a summary by extracting
an optimal chain of scenes via a graph-based approach centered around
the main characters.  In a similar fashion,
\newcite{tsoneva2007automated} create video summaries for TV series
episodes; their algorithm ranks sub-scenes in terms of importance
using features based on character graphs and textual cues available in
the subtitles and movie scripts.  \newcite{vicol2018moviegraphs}
introduce the MovieGraphs dataset, which also uses character-centered
graphs to describe the content of movie video clips.

Our work synthesizes various strands of research on narrative
structure analysis \cite{cutting2016narrative,Hauge:2017}, screenplay
summarization \cite{gorinski-lapata-2015-movie}, and neural network
modeling \cite{Dong2018ASO}. We focus on extractive summarization and
our goal is to identify an optimal sequence of key events in a
narrative. We aim to create summaries which re-tell the plot of a
story in a concise manner.  Inspired by recent neural network-based
approaches
\cite{cheng2016neural,nallapati2017summarunner,zhou2018neural,zheng2019sentence},
we develop supervised and unsupervised models for our summarization
task based on neural representations of scenes and how these relate to
the screenplay's narrative structure. Contrary to most previous work
which has focused on characters, we select summary scenes based on
events and their importance in the story.  Our definition of narrative
structure closely follows \citet{papalampidi2019movie}. However, the
model architectures we propose are general and could be adapted to
different plot analysis schemes \cite{Field:2005,Vogler:2007}. To
overcome the difficulties in evaluating summaries for longer
narratives, we also release a corpus of screenplays with scenes
labeled as important (summary worthy). Our annotations augment an
existing dataset based on CSI episodes \cite{frermann2018whodunnit},
which was originally developed for incremental natural language
understanding.

\section{Problem Formulation}

Let~$\mathcal{D}$ denote a screenplay consisting of a sequence of
scenes $\mathcal{D}=\{s_1, s_2, \dots, s_n\}$. Our aim is to select a
subset $\mathcal{D}' = \{s_i,\dots,s_k\}$ consisting of the most
\emph{informative} scenes (where $k<n$). Note that this definition
produces extractive summaries; we further assume that selected scenes
are presented according to their order in the screenplay. We next
discuss how summaries can be created using both unsupervised and
supervised approaches, and then move on to explain how these are
adapted to incorporate narrative structure.

\subsection{Unsupervised Screenplay
  Summarization} \label{sec:textrank_vanilla}

Our unsupervised model is based on an extension of \textrank
\cite{mihalcea2004textrank, zheng2019sentence}, a well-known algorithm
for extractive single-document summarization.  In our setting, a
screenplay is represented as a graph, in which nodes correspond to
scenes and edges between scenes $s_i$ and $s_j$ are weighted by their
similarity~$e_{ij}$. A node's centrality (importance) is measured by
computing its degree:
\begin{gather}
    \textit{centrality}(s_i) = \lambda_1  \sum_{j<i}e_{ij} + \lambda_2  \sum_{j>i}e_{ij} \label{eq:directed_textrank}
\end{gather}
where $\lambda_1 + \lambda_2 = 1$. The modification introduced in
\citet{zheng2019sentence} takes directed edges into account, capturing
the intuition that the centrality of any two nodes is influenced by
their relative position. Also note that the edges of preceding and
following scenes are differentially weighted by $\lambda_1$
and~$\lambda_2$.

Although earlier implementations of \textrank
\cite{mihalcea2004textrank} compute node similarity based on symbolic
representations such as tf*idf, we adopt a neural
approach. Specifically, we obtain sentence representations based on a
pre-trained encoder. In our experiments, we rely on the Universal
Sentence Encoder (USE; \citealt{cer2018universal}), however, other
embeddings are possible.\footnote{USE performed better than BERT in
  our experiments.}  We represent a scene by the mean of its sentence
representations and measure scene similarity~$e_{ij}$ using
cosine.\footnote{We found cosine to be particularly effective with USE
  representations; other metrics are also possible.} As in the
original \textrank~algorithm \cite{mihalcea2004textrank}, scenes are
ranked based on their centrality and the $M$~most central ones are
selected to appear in the summary.

\subsection{Supervised Screenplay
  Summarization} \label{sec:vanilla_supervised} 

Most extractive models frame summarization as a classification
problem. Following a recent approach (\summarunner; \citealt{nallapati2017summarunner}),
we use a neural network-based encoder to build representations for
scenes and apply a binary classifier over these to predict whether
they should be in the summary. For each scene $s_i \in {\mathcal{D}}$,
we predict a label $y_i \in \{0,1\}$ (where 1 means that $s_i$ must be
in the summary) and assign a score $p(y_i |s_i, \mathcal{D}, \theta)$
quantifying $s_i$'s relevance to the summary ($\theta$ denotes model
parameters). We assemble a summary by selecting~$M$ sentences with the
top~$p(1|s_i, \mathcal{D}, \theta)$.

We calculate sentence representations via the pre-trained USE encoder
\cite{cer2018universal}; a scene is represented as the weighted sum of
the representations of its sentences, which we obtain from a BiLSTM
equipped with an attention mechanism. Next, we compute richer scene
representations by modeling surrounding context of a given scene. We encode the
screenplay with a BiLSTM network and obtain contextualized
representations~$s_i'$ for scenes~$s_i$ by concatenating the hidden
layers of the forward~$\overrightarrow{h_i}$ and
backward~$\overleftarrow{h_i}$ LSTM, respectively:
$ s_i' = [\overrightarrow{h_i} ; \overleftarrow{h_i}]$.  The vector
$s_i'$ therefore represents the \textit{content} of the $i^{th}$
scene.

We also estimate the \textit{salience} of scene~$s_i$ by measuring its
similarity with a global screenplay content representation~$d$. The
latter is the weighted sum of all scene representations
$s_1, s_2, \dots, s_n$. We calculate the semantic similarity
between~$s_i'$ and~$d$ by computing the element-wise dot product~$b_i$,
cosine similarity~$c_i$, and pairwise distance~$u_i$ between
their respective vectors:
\begin{gather}
b_i = s_i' \odot d \quad 
    c_i = \frac{s_i' \cdot d}{\norm{s_i'}\norm{d}} \label{eq:interaction2} \\
    u_i = \dfrac{s_i' \cdot d}{\max(\Vert s_i' \Vert _2 \cdot \Vert d
      \Vert _2)}\label{eq:interaction3} 
\end{gather}
The salience $v_i$ of scene $s_i$ is the concatenation of the
similarity metrics: $v_i = [b_i ; c_i ; u_i]$.  The content
vector~$s_i'$ and the salience vector~$v_i$ are concatenated and fed
to a single neuron that outputs the probability of a scene belonging
to the summary.\footnote{Aside from salience and content,
  \citet{nallapati2017summarunner} take into account novelty and
  position-related features. We ignore these as they are specific to
  news articles and denote the modified model as \summarunnerdif.}

\subsection{Narrative Structure} 
\label{sec:narrative_structure}

We now explain how to inject knowledge about narrative structure into
our summarization models. For both models, such knowledge is
transferred via a network pre-trained on the
TRIPOD\footnote{\url{https://github.com/ppapalampidi/TRIPOD}} dataset
introduced by \newcite{papalampidi2019movie}. This dataset contains 99
movies annotated with turning points. TPs are key events in a
narrative that define the progression of the plot and occur between
consecutive acts (thematic units). It is often assumed
\cite{cutting2016narrative} that there are six acts in a film
(Figure~\ref{fig:tps_example}), each delineated by a turning point
(arrows in the figure). Each of the five TPs has also a well-defined
function in the narrative: we present each TP alongside with its
definition as stated in screenwriting theory \cite{Hauge:2017} and
adopted by \newcite{papalampidi2019movie} in
Table~\ref{tab:tp_description} (see Appendix~\ref{app:theory} for a
more detailed description of narrative structure theory).
\newcite{papalampidi2019movie} identify scenes in movies that
correspond to these key events as a means for analyzing the narrative
structure of movies. They collect sentence-level TP annotations for
plot synopses and subsequently project them via distant supervision
onto screenplays, thereby creating silver-standard labels. We utilize
this silver-standard dataset in order to pre-train a network which
performs TP identification.

\paragraph{TP Identification Network}
We first encode screenplay scenes via a BiLSTM equipped with an
attention mechanism. We then contextualize them with respect to the
whole screenplay via a second BiLSTM. Next, we compute topic-aware
scene representations~$t_{i}$ via a context interaction layer (CIL) as
proposed in \newcite{papalampidi2019movie}. 
CIL is inspired by traditional segmentation approaches
\cite{hearst1997texttiling} and measures the semantic similarity of
the current scene with a preceding and following context window in the
screenplay. Hence, the topic-aware scene representations also encode
the degree to which each scene acts as a topic boundary in the screenplay.

In the final layer, we
employ TP-specific attention mechanisms to compute the probability
$p_{ij}$ that scene $t_i$ represents the $j^{th}$ TP in the
screenplay. Note that we expect the TP-specific attention
distributions to be sparse, as there are only a few scenes which are
relevant for a TP (recall that TPs are boundary scenes between
sections). To encourage sparsity, we add a low temperature value $\tau$
\cite{hinton2015distilling} to the softmax part of the attention
mechanisms:
\begin{gather}
    g_{ij} = \tanh(W_j t_i + b_j)\label{eq:att_ei_ent}, \quad g_j \in [-1,1] \\
p_{ij} = \dfrac{\exp(g_{ij}/\tau)}{\sum_{t=1}^{T} \exp(g_{tj}/\tau)} \label{eq:sharp_p}, \quad \sum_{i=1}^{T} p_{ij} = 1 
\end{gather}
where $W_j,b_j$ represent the trainable weights of the attention layer
of the $j^{th}$~TP.

\begin{table}[t]
\small
\centering
\begin{tabular}{@{}p{0.4\columnwidth}@{~~}p{0.6\columnwidth}@{}}
\hline
\multicolumn{1}{c}{{Turning Point}} & \multicolumn{1}{c}{{Definition}} \\ \hline
{\color{tp1}{TP1: Opportunity}} & Introductory event that occurs after the presentation of the story setting. \\ \hdashline
{\color{tp2}{TP2: Change of Plans}} & Event where the main goal of the story is defined. \\ \hdashline
{\color{tp3}{TP3: Point of No Return}} & Event that pushes the main character(s) to fully commit to their goal. \\ \hdashline
{\color{tp4}{TP4: Major Setback}} & Event where everything falls apart (temporarily or permanently). \\ \hdashline
{\color{tp5}{TP5: Climax}} & Final event of the main story, moment of resolution. \\ \hline
\end{tabular}
\caption{Turning points and their definitions as given in \newcite{papalampidi2019movie}}
\label{tab:tp_description}
\end{table}

\paragraph{Unsupervised \textsc{\summer}}
We now introduce our model, \summer (short for \ul{S}creenplay
S\ul{umm}arization with Narrativ\ul{e} St\ul{r}ucture).\footnote{We
  make our code publicly available at
  \url{https://github.com/ppapalampidi/SUMMER}.}  We first present an
unsupervised variant which modifies the computation of scene
centrality in the directed version of \textrank
(Equation~\eqref{eq:directed_textrank}).

Specifically, we use the pre-trained network described in
Section~\ref{sec:narrative_structure} to obtain TP-specific attention
distributions.
We then select an overall score~$f_i$ for each scene (denoting how
likely it is to act as a TP). We set~$f_i = \max_{j \in [1,5]}p_{ij}$,
i.e., to the~$p_{ij}$ value that is highest across TPs.  We
incorporate these scores into centrality as follows:
{
\thinmuskip=0mu\medmuskip=0mu\thickmuskip=0mu
\begin{align}
\textit{centrality}(s_i) = \lambda_1 \sum_{j<i}(e_{ij} + f_j)
                         + \lambda_2 \sum_{j>i}(e_{ij} + f_i) \label{eq:our_textrank}
\end{align}%
}%
Intuitively, we add the $f_j$ term in the forward sum in order to incrementally increase the centrality scores of scenes as the story moves on and we encounter more TP events (i.e.,~we move to later sections in the narrative). At the same time, we add the $f_i$ term in the backward sum in order to also increase the scores of scenes identified as TPs.

\begin{figure}[t]
    \centering
    \includegraphics[width=\columnwidth,page=1]{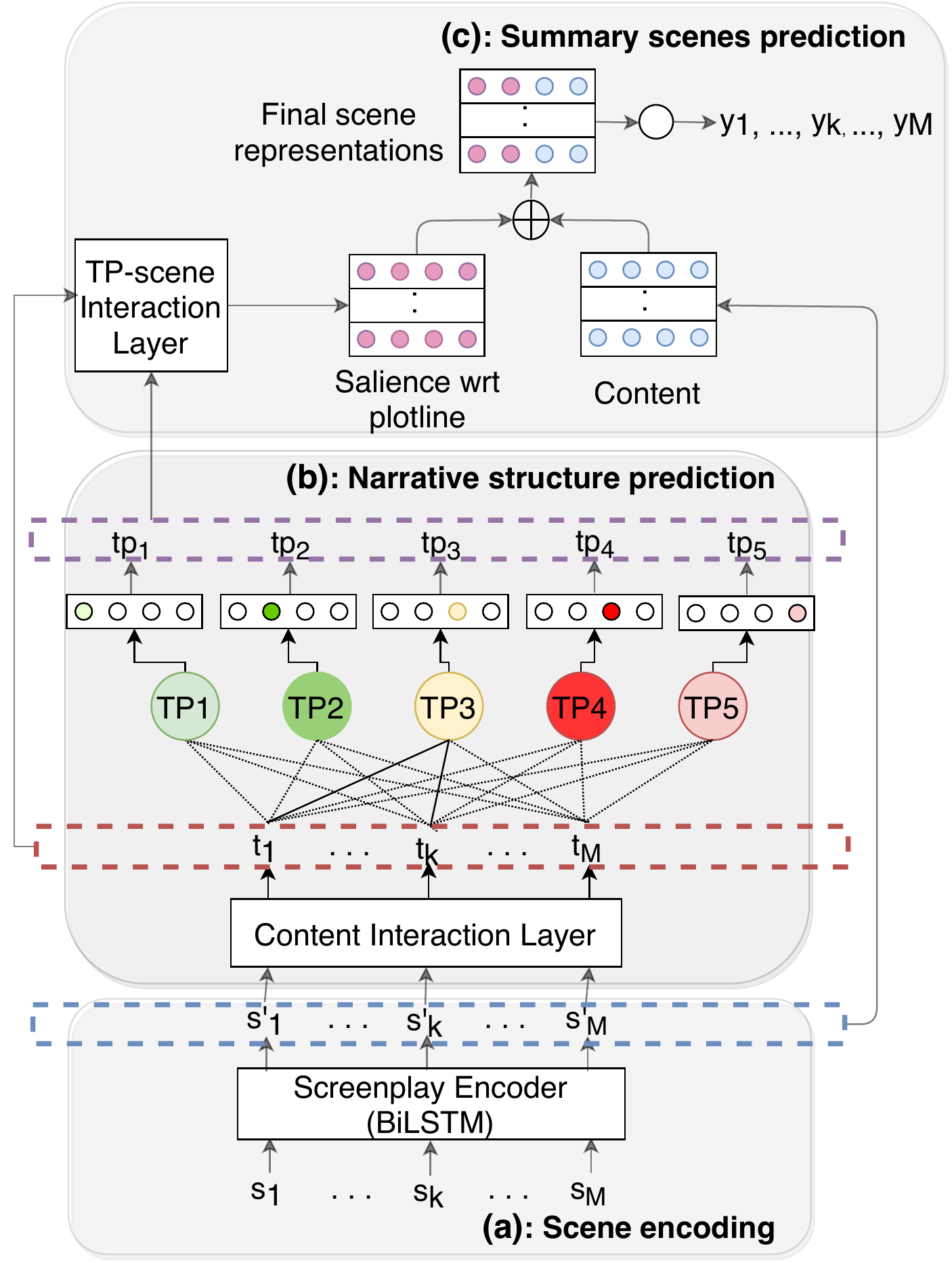}
    \caption{Overview of \summer. We use one \mbox{TP-specific}
      attention mechanism per turning point in order to acquire
      TP-specific distributions over scenes. We then compute the
      similarity between TPs and contextualized scene
      representations. Finally, we perform max pooling over
      TP-specific similarity vectors and concatenate the final
      similarity representation with the contextualized scene
      representation.}
    \label{fig:summer_model}
\end{figure}

\paragraph{Supervised \summer} \label{sec:summer}

We also propose a supervised variant of \summer following the basic
model formulation in Section~\ref{sec:summer}. We still represent a
scene as the concatenation of a content vector~$s'$ and salience
vector~$v'$, which serve as input to a binary classifier. However, we
now modify how salience is determined; instead of computing a general
global content representation~$d$ for the screenplay, we identify a sequence of TPs and measure the semantic similarity of each scene
with this sequence. Our model is depicted in
Figure~\ref{fig:summer_model}.

We utilize the pre-trained TP network
(Figures~\ref{fig:summer_model}(a) and~(b)) to compute sparse
attention scores over scenes. In the supervised setting, where
gold-standard binary labels provide a training signal, we fine-tune
the network in an end-to-end fashion on summarization
(Figure~\ref{fig:summer_model}(c)). 
We compute the TP representations via the attention scores; we calculate a vector~$tp_j$ as the weighted sum of all
topic-aware scene representations $t$ produced via CIL: $tp_j = \sum_{i\in[1,N]}p_{ij}t_i$,
where $N$ is the number of scenes in a screenplay. In practice,
only a few scenes contribute to~$tp_j$ due to the $\tau$~parameter in
the softmax function (Equation~\eqref{eq:sharp_p}).

A TP-scene interaction layer measures the semantic similarity between
scenes~$t_i$ and latent TP representations~$tp_j$
(Figure~\ref{fig:summer_model}(c)).  Intuitively, a complete summary
should contain scenes which are related to at least one of the key
events in the screenplay.  We calculate the semantic
similarity~$v_{ij}$ of scene~$t_i$ with TP~$tp_j$ as in
Equations~\eqref{eq:interaction2} and~\eqref{eq:interaction3}. We then
perform max pooling over vectors $v_{i1}, \dots, v_{iT}$, where~$T$ is
the number of TPs (i.e.,~five) and calculate a final similarity
vector~$v_i'$ for the $i^{th}$ scene.

The model is trained end-to-end on the summarization task using
\textit{BCE}, the binary cross-entropy loss function. We add an
extra regularization term to this objective to encourage the
TP-specific attention distributions to be orthogonal (since we want
each attention layer to attend to different parts of the screenplay).
We thus maximize the Kullback-Leibler (KL)
divergence~$\mathcal{D}_{KL}$ between all pairs of TP attention
distributions $tp_i$, $i \in [1,5]$:
\begin{gather}
        O = \sum_{i \in [1,5]}\sum_{j \in [1,5], j \ne i}\log
        \frac{1}{\mathcal{D}_{KL}\left(tp_i \middle\| tp_j\right) +
          \epsilon}  \label{req:1}
\end{gather}
Furthermore, we know from screenwriting theory \cite{Hauge:2017} that
there are rules of thumb as to when a TP should occur (e.g.,~the
Opportunity occurs after the first 10\% of a screenplay, Change of
Plans is approximately 25\% in). It is reasonable to discourage~$tp$
distributions to deviate drastically from these expected
positions. Focal regularization~$F$ minimizes the KL divergence
$\mathcal{D}_{KL}$ between each TP attention distribution $tp_i$ and
its expected position distribution $th_i$:
\begin{gather}
    F = \sum_{i \in [1,5]}\mathcal{D}_{KL}\left(tp_i \middle\|
      th_i\right) \label{reg:2}
\end{gather}
The final loss $\mathcal{L}$ is the weighted sum of all three
components, where $a,b$ are fixed during training: $\mathcal{L} = \textit{BCE} + a O + b F$.

\section{Experimental Setup}

\paragraph{Crime Scene Investigation Dataset}
We performed experiments on an extension of the CSI dataset\footnote{\url{https://github.com/EdinburghNLP/csi-corpus}} introduced
by \newcite{frermann2018whodunnit}. It consists of 39~CSI episodes,
each annotated with \emph{word-level} labels denoting whether the
perpetrator is mentioned in the utterances characters speak. We
further collected \emph{scene-level} binary labels indicating whether
episode scenes are important and should be included in a summary.
Three human judges performed the annotation task after watching the
CSI episodes scene-by-scene. To facilitate the annotation, judges were
asked to indicate why they thought a scene was important, citing the
following reasons: it revealed
\begin{enumerate*}[label=(\roman*)]
    \item the victim,
    \item the cause of death,
    \item an autopsy report,
    \item crucial evidence,
    \item the perpetrator, and
    \item the motive or the relation between perpetrator and victim.
\end{enumerate*}
Annotators were free to select more than one or none of the listed
reasons where appropriate.  We can think of these reasons as
high-level \emph{aspects} a good summary should cover (for CSI and
related crime series). Annotators were not given any information about
TPs or narrative structure; the annotation was not guided by
theoretical considerations, rather our aim was to produce useful CSI
summaries.  Table~\ref{tab:csi_stats} presents the dataset statistics
(see also Appendix~\ref{app:corpus} for more detail).

\begin{table}[t]
\small
\centering
\begin{tabular}{L{12em}R{7em}}\hline
\multicolumn{2}{c}{\emph{overall}} \\\hline
episodes & 39 \\
scenes & 1544 \\ 
summary scenes & 454 \\ 
\hline
\multicolumn{2}{c}{\textit{per episode}} \\ \hline
scenes & 39.58 (6.52) \\
crime-specific aspects & 5.62 (0.24) \\
summary scenes & 11.64 (2.98) \\
summary scenes (\%) & 29.75 (7.35) \\ 
sentences & 822.56 (936.23) \\
tokens & 13.27k (14.67k) \\ \hline
\multicolumn{2}{c}{\textit{per episode scene}} \\ \hline
sentences & 20.78 (35.61) \\
tokens & 335.19 (547.61) \\
tokens per sentence & 16.13 (16.32) \\
 \hline
\end{tabular}
\caption{CSI dataset statistics; means and (std).} 
\label{tab:csi_stats}
\end{table}

\paragraph{Implementation Details}

In order to set the hyperparameters of all proposed networks, we used
a small development set of four episodes from the CSI dataset (see Appendix~\ref{app:corpus} for details). After
experimentation, we set the temperature~$\tau$ of the softmax layers
for the TP-specific attentions (Equation~\eqref{eq:sharp_p})
to~0.01. Since the binary labels in the supervised setting are
imbalanced, we apply class weights to the binary cross-entropy loss of
the respective models. We weight each class by its inverse frequency
in the training set. Finally, in supervised \summer, where we also
identify the narrative structure of the screenplays, we consider as
key events per TP the scenes that correspond to an attention score
higher than~0.05. More implementation details can be found in
Appendix~\ref{app:implementation}.

As shown in Table~\ref{tab:csi_stats}, the gold-standard summaries in
our dataset have a compression rate of approximately 30\%. During
inference, we select the top $M$~scenes as the summary, such that they
correspond to 30\% of the length of the episode.

\section{Results and Analysis}
\label{sec:results}

\paragraph{Is Narrative Structure Helpful?}
We perform 10-fold cross-validation and evaluate model performance in
terms of F1~score. Table~\ref{tab:unsupervised_results} summarizes the
results of unsupervised models.  We present the following baselines:
{Lead~30\%} selects the first~30\% of an episode as the summary,
{Last~30\%} selects the last~30\%, and {Mixed~30\%}, randomly selects
15\%~of the summary from the first~30\% of an episode and~15\% from
the last~30\%.  We also compare \summer against \textrank based on
tf*idf \cite{mihalcea2004textrank}, the directed neural variant
described in Section~\ref{sec:textrank_vanilla} without any TP
information, a variant where TPs are approximated by their expected
position as postulated in screenwriting theory, and a variant that
incorporates information about characters
\cite{gorinski-lapata-2015-movie} instead of narrative structure. For
the character-based \textrank, called \scenesum, we substitute the
$f_i, f_j$ scores in Equation~\eqref{eq:our_textrank} with
character-related importance scores $c_i$ similar to the definition in
\newcite{gorinski-lapata-2015-movie}:
\begin{gather}
    c_i = \frac{\sum_{c \in C}\ [c \in S\ \cup\
      \textit{main}(C)]}{\sum_{c \in C}\ [c \in S]}
 \label{eq:char_scores}
\end{gather}
where $S$ is the set of all characters participating in scene $s_i$,
$C$ is the set of all characters participating in the screenplay and
$\textit{main}(C)$ are all the main characters of the screenplay. We
retrieve the set of main characters from the IMDb page of the
respective episode.  We also note that human agreement for our task
is 79.26 F1~score, as measured on a small subset of the corpus.

\begin{table}[t]
\small
    \centering
    \begin{tabular}{lc}
    \hline
     Model &  F1 \\ \hline
    Lead 30\% & 30.66 \\
    Last 30\% & 39.85 \\
    \begin{tabular}[c]{@{}l@{}}Mixed~30\%\end{tabular} & 34.32 \\ \hline
    \textrank, undirected, tf*idf & 32.11\\
\textrank,     directed, neural &  41.75 \\ 
\textrank,    directed,   expected TP positions  & 41.05 \\
\scenesum,     directed, character-based weights &  42.02 \\ 
    \summer& \textbf{44.70} \\

    \hline
    \end{tabular}
    \caption{Unsupervised screenplay summarization.}
    \label{tab:unsupervised_results}
\end{table}

As shown in Table~\ref{tab:unsupervised_results}, \summer achieves the
best performance (44.70 F1~score) among all models and is superior to
an equivalent model which uses expected TP positions or a
character-based representation. This indicates that the pre-trained
network provides better predictions for key events than position and
character heuristics, even though there is a domain shift from
Hollywood movies in the TRIPOD corpus to episodes of a crime series in
the CSI corpus. Moreover, we find that the directed versions of
\textrank are better at identifying important scenes than the
undirected version. We found that performance peaks
with~\mbox{$\lambda_1= 0.7$} (see Equation~\eqref{eq:our_textrank}),
indicating that higher importance is given to scenes as the story
progresses (see Appendix~\ref{app:further_results} for experiments
with different~$\lambda$ values).

\begin{table}[t]
\small
\centering
\begin{tabular}{@{}l@{~~~}c@{~~~}c@{~~~}c@{}}
\hline
&   F1  & \begin{tabular}[c]{@{}c@{}} Coverage  \\ of aspects  \end{tabular} & \begin{tabular}[c]{@{}l@{}} \# scenes \\ per TP \end{tabular} \\ \hline
Lead 30\%   & 30.66 & -- & -- \\
Last 30\%   & 39.85 & -- & -- \\
Mixed 30\%   & 34.32 & -- & --\\
\hline
\summarunnerdif  & 48.56  & -- & -- \\ 
\scenesum  & 47.71  & -- & -- \\ 
\summer, fixed one-hot TPs &
46.92 & 63.11  & 1.00 \\
\summer,  {fixed distributions} & 47.64 & 67.01 &
 1.05\\
\summer, $-$P, $-$R & \textbf{51.93} & 44.48  & 1.19 \\
\summer, $-$P, $+$R & 49.98 & 51.96  & 1.14  \\
\summer, $+$P, $-$R  & {50.56} & 62.35  & 3.07 \\
\summer, $+$P, $+$R  & \textbf{52.00} & \textbf{70.25}  & 1.20  \\ 
\hline
\end{tabular}
\caption{Supervised screenplay summarization; for  in
  \summer~variants, we  also report the percentage of aspect labels
  covered by latent TP predictions.}
\label{tab:results_generic}
\end{table}

In Table~\ref{tab:results_generic}, we report results for supervised
models. Aside from the various baselines in the first block of the
table, we compare the neural extractive model
\summarunnerdif\footnote{Our adaptation of \summarunner that considers
  content and salience vectors for scene selection.}
\cite{nallapati2017summarunner} presented in
Section~\ref{sec:vanilla_supervised} with several variants of our
model \summer.  We experimented with randomly initializing the network
for TP identification ($-$P) and with using a pre-trained network
($+$P). We also experimented with removing the regularization terms,
$O$ and $F$ (Equations~(\ref{req:1}) and (\ref{reg:2})) from the loss
($-$R). We assess the performance of \summer when we follow a two-step
approach where we first predict TPs via the pre-trained network and
then train a network on screenplay summarization based on fixed TP
representations ({fixed one-hot TPs}), or alternatively use expected
TP position distributions as postulated in screenwriting theory
({fixed distributions}). Finally, we incorporate character-based
information into our baseline and create a supervised version of
\scenesum. We now utilize the character importance scores per scene
(Equation~\eqref{eq:char_scores}) as attention scores -- instead of
using a trainable attention mechanism -- when computing the global
screenplay representation $d$ (Section~\ref{sec:vanilla_supervised}).

Table~\ref{tab:results_generic} shows that all end-to-end \summer variants
outperform \summarunnerdif. The best result (52.00 F1~Score) is
achieved by pre-trained \summer with regularization, outperforming
\summarunnerdif by an absolute difference of~3.44. The randomly
initialized version with no regularization achieves similar
performance (51.93 F1~score).  For summarizing screenplays, explicitly
encoding narrative structure seems to be more beneficial than general
representations of scene importance. Finally, two-step versions of
\summer perform poorly, which indicates that end-to-end training and
fine-tuning of the TP identification network on the target dataset
is crucial.

\begin{table*}[t]
\small
\centering
\begin{tabular}{L{16em}rrrrrr}
\hline
Turning Point & Crime scene & Victim  & Death Cause & Perpetrator & Evidence  & Motive \\ \hline
\begin{tabular}[c]{@{}l@{}} Opportunity \end{tabular} & \textbf{56.76} & \textbf{52.63} & 15.63 & 15.38  & 2.56 & 0.00 \\
\begin{tabular}[c]{@{}l@{}} Change of Plans \end{tabular} & \textbf{27.03} & \textbf{42.11} & \textbf{21.88} & 15.38  & 5.13 & 0.00 \\ 
\begin{tabular}[c]{@{}l@{}} Point of no Return \end{tabular} & 8.11 & 13.16  & 9.38 & \textbf{25.64}  & \textbf{48.72} & 5.88 \\
\begin{tabular}[c]{@{}l@{}} Major Setback \end{tabular} & 0.00 & 0.00 & 6.25 & 10.25  & \textbf{48.72} & \textbf{35.29} \\
\begin{tabular}[c]{@{}l@{}} Climax \end{tabular} & 2.70 & 0.00 & 6.25 & 2.56  & \textbf{23.08} & \textbf{55.88} \\  \hline
\end{tabular}
\caption{Percentage of aspect labels covered per TP for \summer, $+$P, $+$R.}
\label{tab:TPs_aspects}
\end{table*}

\begin{table*}[t]
\small
\centering
\begin{tabular}{lrrrrrrrr}
\hline
System & Crime scene & Victim & Death Cause & Perpetrator &
Evidence & Motive & Overall & Rank\\ \hline
\begin{tabular}[c]{@{}l@{}} \summarunnerdif \end{tabular} & 85.71 &
\textbf{93.88} &  75.51 & 81.63 & 59.18 & 38.78 & 72.45 & 2.18\\ 
\begin{tabular}[c]{@{}l@{}} \summer \end{tabular} & \textbf{89.80} &
87.76 & \textbf{83.67} & 81.63  & \textbf{77.55} & \textbf{57.14} &
\textbf{79.59} & 2.00 \\ 
\begin{tabular}[c]{@{}l@{}} Gold \end{tabular} & \textbf{89.80} &
91.84 &  71.43 & \textbf{83.67}  & 65.31 & \textbf{57.14} & 76.53 & 1.82\\
\hline
\end{tabular}
\caption{Human evaluation: percentage of yes answers by AMT workers
  regarding each aspect in a summary. All differences in (average)
Rank are significant ($p<0.05$, using a $\chi^{2}$ test).}
\label{tab:human_evaluation_per_aspect}
\end{table*}

\paragraph{What Does the Model Learn?} 
Apart from performance on summarization, we would also like
to examine the quality of the TPs inferred by \summer (supervised
variant). Problematically, we do not have any gold-standard TP
annotation in the CSI corpus. Nevertheless, we can implicitly assess
whether they are meaningful by measuring how well they correlate with
the reasons annotators cite to justify their decision to include a
scene in the summary (e.g.,~because it reveals cause of death or
provides important evidence).  Specifically, we compute the extent to
which these aspects overlap with the TPs predicted by \summer as:
{
\thinmuskip=0mu\medmuskip=0mu\thickmuskip=0mu
\begin{gather}
    C = \frac{\sum_{A_i \in A} \sum_{TP_j \in TP}\ [\textit{dist}(TP_j,A_i) \leq 1]}{|A|}
\label{eq:overlap}%
\end{gather}%
}
where $A$ is the set of all aspect scenes, $|A|$ is the number of aspects,  $TP$ is the set of scenes
inferred as TPs by the model, 
$A_i$ and $TP_j$ are the subsets of
scenes corresponding to the $i^{th}$ aspect and $j^{th}$ TP,
respectively, and~$\textit{dist}(TP_j,A_i)$ is the minimum distance
between~$TP_j$ and $A_i$~in number of scenes.

The proportion of aspects covered is given in
Table~\ref{tab:results_generic}, middle column. We find that
coverage is relatively low (44.48\%) for the randomly initialized \summer~with no regularization. There is a slight
improvement of 7.48\% when we force the TP-specific attention
distributions to be orthogonal and close to expected
positions. Pre-training and regularization provide a significant
boost, increasing coverage to~70.25\%, while pre-trained
\summer~without regularization infers on average more scenes
representative of each TP. This shows that the orthogonal constraint also encourages sparse attention distributions
for~TPs.

Table~\ref{tab:TPs_aspects} shows the degree of association between
individual TPs and summary aspects (see Appendix~\ref{app:further_results} for illustrated examples). We observe that
Opportunity and Change of Plans are mostly associated with
information about the crime scene and the victim, Climax is
focused on the revelation of the motive, while information relating to
cause of death, perpetrator, and evidence is captured by both
Point of no Return and Major Setback. 
Overall, the generic Hollywood-inspired TP labels are adjusted to our genre and describe crime-related key events, even though no aspect labels were provided to our model during training.

\paragraph{Do Humans Like the Summaries?} We also conducted a human
evaluation experiment using the summaries created for 10~CSI
episodes.\footnote{\url{https://github.com/ppapalampidi/SUMMER/tree/master/video_summaries}}
We produced summaries based on the gold-standard annotations (Gold),
\summarunnerdif, and the supervised version of \summer. Since 30\%~of
an episode results in lengthy summaries (15 minutes on average), we
further increased the compression rate for this experiment by limiting
each summary to six scenes. For the gold standard condition, we
randomly selected exactly one scene per aspect. For \summarunnerdif
and \summer we selected the top six predicted scenes based on their
posterior probabilities. We then created video summaries by isolating
and merging the selected scenes in the raw video.

We asked Amazon Mechanical Turk (AMT) workers to watch the video
summaries for all systems and rank them from most to least
informative. They were also presented with six questions relating to
the aspects the summary was supposed to cover (e.g.,~Was the victim
revealed in the summary? Do you know who the perpetrator was?). They
could answer Yes, No, or Unsure. Five workers evaluated each summary.

Table~\ref{tab:human_evaluation_per_aspect} shows the proportion of
times participants responded Yes for each aspect across the three
systems.  Although \summer does not improve over \summarunnerdif in
identifying basic information (i.e.,~about the victim and
perpetrator), it creates better summaries overall with more diverse
content (i.e.,~it more frequently includes information about cause of
death, evidence, and motive). This observation validates our
assumption that identifying scenes that are semantically close to the
key events of a screenplay leads to more complete and detailed
summaries.  Finally, Table~\ref{tab:human_evaluation_per_aspect} also
lists the average rank per system (lower is better), which shows that
crowdworkers like gold summaries best, \summer is often ranked second,
followed by \summarunnerdif in third place.

\section{Conclusions}
\label{sec:conclusions}

In this paper we argued that the underlying structure of narratives is
beneficial for long-form summarization. We adapted a scheme for
identifying narrative structure (i.e.,~turning points) in Hollywood
movies and showed how this information can be integrated with
supervised and unsupervised extractive summarization
algorithms. Experiments on the CSI corpus showed that this scheme
transfers well to a different genre (crime investigation) and that
utilizing narrative structure boosts summarization performance,
leading to more complete and diverse summaries. Analysis of model
output further revealed that latent events encapsulated by turning
points correlate with important aspects of a CSI summary.

Although currently our approach relies solely on textual information,
it would be interesting to incorporate additional modalities such as
video or audio. Audiovisual information could facilitate the
identification of key events and scenes. Besides narrative structure,
we would also like to examine the role of \emph{emotional arcs}
\cite{Vonnegut:1981,Reagan:ea:2016} in a screenplay. An often integral
part of a compelling story is the emotional experience that is evoked
in the reader or viewer (e.g., somebody gets into trouble and then out
of it, somebody finds something wonderful, loses it, and then finds it
again). Understanding emotional arcs may be useful to revealing a
story's shape, highlighting important scenes, and tracking how the
story develops for different characters over time.

\section*{Acknowledgments}
We thank the anonymous reviewers for their feedback.  We gratefully
acknowledge the support of the European Research Council (Lapata;
award 681760, ``Translating Multiple Modalities into Text'') and of
the Leverhulme Trust (Keller; award IAF-2017-019).

\bibliography{acl2020}
\bibliographystyle{acl_natbib}

\appendix

\section{Narrative Structure Theory}
\label{app:theory}

The initial formulation of narrative structure was promoted by Aristotle, who defined the basic triangle-shaped plot structure, that has a beginning (\textit{protasis}), middle (\textit{epitasis}) and end (\textit{catastrophe}) \cite{pavis1998dictionary}. 
However, later theories argued that the structure of a play should be more complex \cite{brink2011horace} and hence, other schemes \cite{freytag1896freytag} were proposed with fine-grained stages and events defining the progression of the plot. 
 These events are considered as the precursor of turning points, defined by \newcite{thompson1999storytelling} and used in modern variations of screenplay theory. 
Turning points are narrative moments from which the plot goes in a different direction. By definition these occur at the junctions of acts. 

Currently, there are myriad schemes describing the narrative structure
of films, which are often used as a practical guide for screenwriters
\cite{cutting2016narrative}. One variation of these modern schemes is
adopted by \newcite{papalampidi2019movie}, who focus on the definition
of turning points and demonstrate that such events indeed exist in
films and can be automatically identified.  According to the adopted
scheme \cite{Hauge:2017}, there are six stages (acts) in a film,
namely \textit{the setup}, \textit{the new situation},
\textit{progress}, \textit{complications and higher stakes},
\textit{the final push} and \textit{the aftermath}, separated by the
five turning points presented in Table \ref{tab:tp_description}.

\begin{figure}[t]
    \centering
    \vspace{-1.7em}
    \includegraphics[width=\columnwidth]{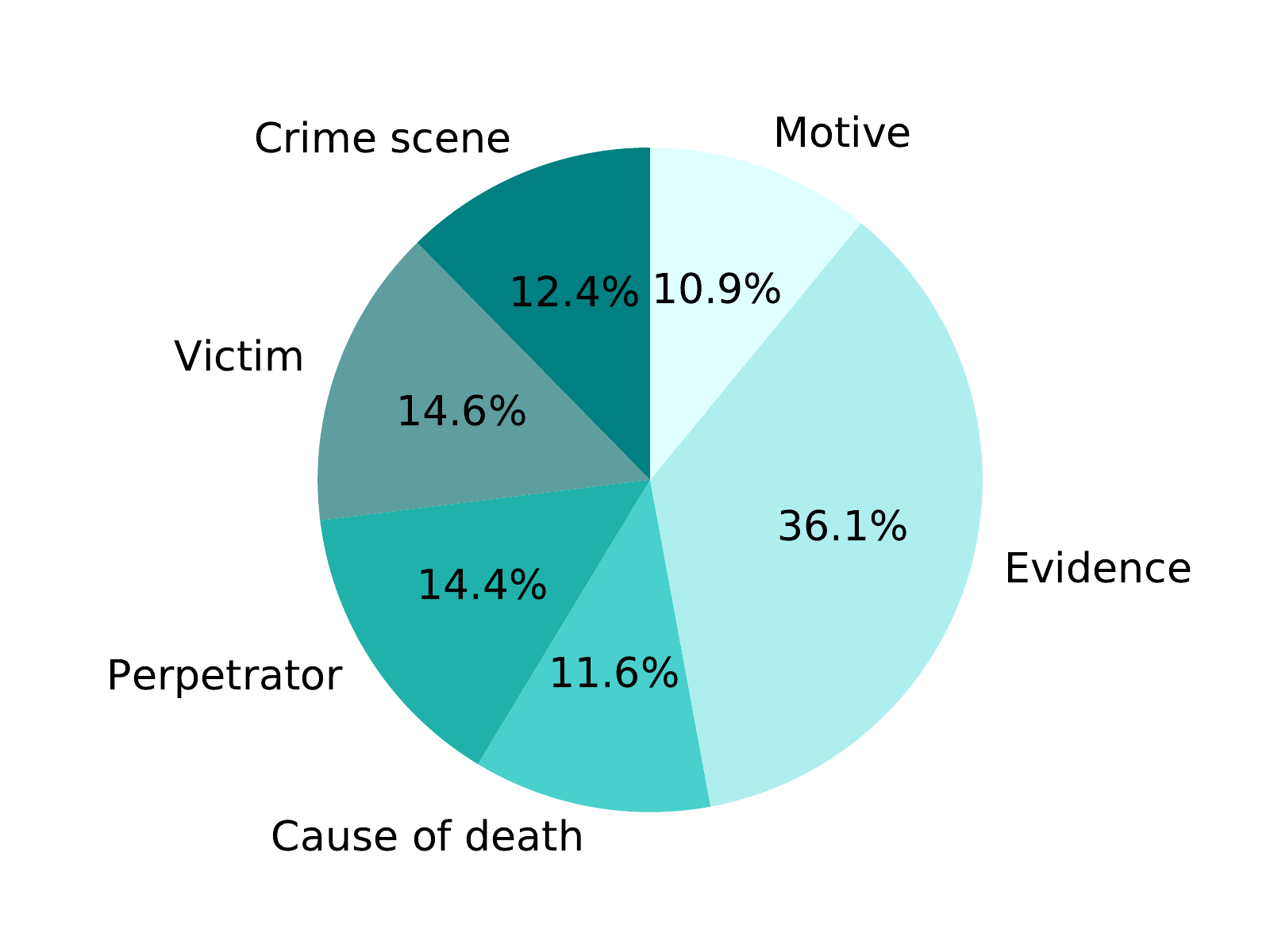}
    \caption{Average composition of a CSI summary based on  different crime-related aspects.}
    \label{fig:distribution_aspects}
\end{figure}

\section{CSI Corpus}
\label{app:corpus}

As described in Section 4, we collected aspect-based summary labels
for all episodes in the CSI corpus. In
Figure~\ref{fig:distribution_aspects} we illustrate the average
composition of a summary based on the different aspects seen in a
crime investigation (e.g.,~crime scene, victim, cause of death,
perpetrator, evidence). Most of these aspects are covered in
\mbox{10--15\%} of a summary, which corresponds to approximately two
scenes in the episode. Only the ``Evidence'' aspect occupies a larger
proportion of the summary (36.1\%) corresponding to five
scenes. However, there exist scenes which cover multiple aspects (an
as a result are annotated with more than one label) and episodes that
do not include any scenes related to a specific aspect (e.g.,~if the
murder was a suicide, there is no perpetrator).

We should note that \newcite{frermann2018whodunnit} discriminate
between different cases presented in the same episode in the original
CSI dataset. Specifically, there are episodes in the dataset, where
except for the primary crime investigation case, a second one is
presented occupying a significantly smaller part of the
episode. Although in the original dataset, there are annotations
available indicating which scenes refer to each case, we assume no
such knowledge treating the screenplay as a single unit --- most TV
series and movies contain sub-stories. We also hypothesize that the
latent identified TP events in \summer should relate to the primary
case.

\section{Implementation Details}
\label{app:implementation}
\begin{figure}[t]
    \centering
    \vspace{-1.7em}
    \includegraphics[width=\columnwidth]{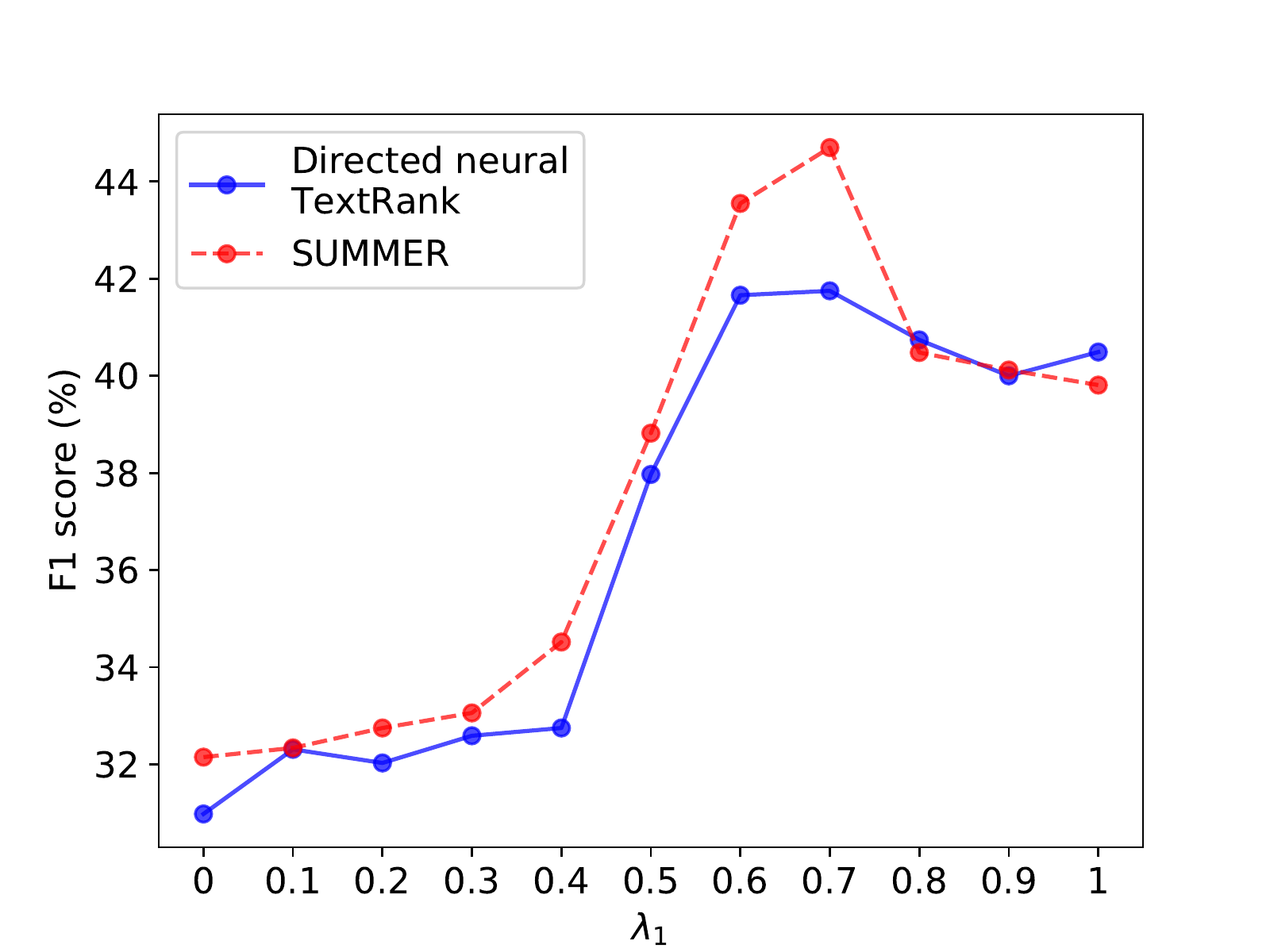}
    \caption{F1 score (\%) for directed neural \textrank~and \summer
      for unsupervised summarization with respect to different
      $\lambda_1$ values. Higher $\lambda_1$ values correspond to
      higher importance in the next context for the centrality
      computation of a current scene.}
    \label{fig:unsupervised_f1_lambda}
\end{figure}

\begin{figure*}[t]
    \centering
\hspace*{-1.7cm}\includegraphics[width=19cm]{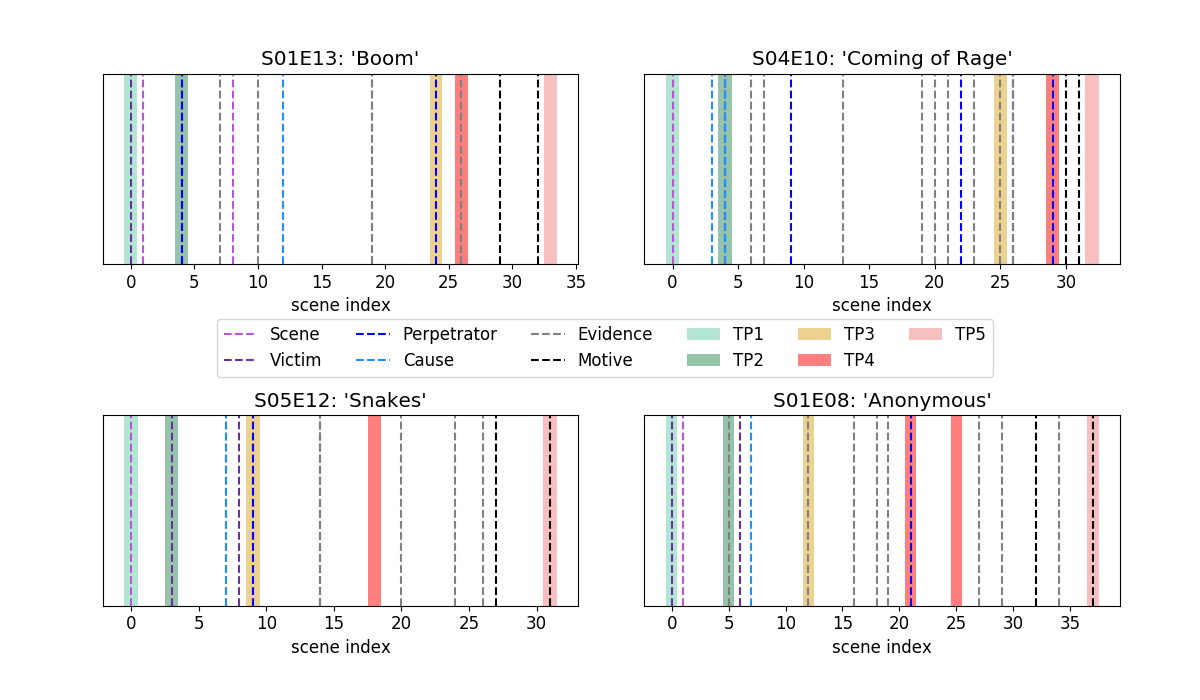}
\caption{Examples of inferred TPs alongside with gold-standard
  aspect-based summary labels in CSI episodes at test time. The TP
  events are identified in the latent space for the supervised version
  of \summer ($+$P, $+$R).}
    \label{fig:TP_examples}
\end{figure*}

In all unsupervised versions of \textrank and \summer we used a
threshold $h$ equal to 0.2 for removing weak edges from the
corresponding fully connected screenplay graphs. For the supervised
version of \summer, where we use additional regularization terms in
the loss function, we experimentally set the weights $a$ and $b$ for
the different terms to 0.15 and 0.1, respectively.

We used the Adam algorithm \cite{kingma2014adam} for optimizing our
networks. After experimentation, we chose an LSTM 
with 64 neurons for encoding the scenes in the screenplay and another identical one for contextualizing them. For the context
interaction layer, the window~$l$ for computing the surrounding context of a screenplay scene was set to 20\% of the screenplay length as proposed in \newcite{papalampidi2019movie}. Finally, we also added a dropout of~0.2. For developing our models we used PyTorch
\cite{paszke2017automatic}.

\section{Additional Results}
\label{app:further_results}

We illustrate in Figure \ref{fig:unsupervised_f1_lambda} the
performance (F1 score) of the directed neural \textrank and \summer
models in the unsupervised setting with respect to different
$\lambda_1$ values. Higher $\lambda_1$ values correspond to higher
importance for the succeeding scenes and respectively lower importance
for the preceding ones, since $\lambda_1$ and $\lambda_2$ are bounded
($\lambda_1 + \lambda_2 = 1$).

We observe that performance increases when higher importance is
attributed to screenplay scenes as the story moves on
($\lambda_1>0.5$), whereas for extreme cases ($\lambda_1 \to 1$),
where only the later parts of the story are considered, performance
drops. Overall, the same peak appears for both \textrank and \summer
when $\lambda_1 \in [0.6,0.7]$, which means that slightly higher
importance is attributed to the screenplay scenes that follow.
Intuitively, initial scenes of an episode tend to have high similarity with all other scenes in the screenplay, and on their own are not very informative (e.g., the crime, victim, and suspects are introduced but the perpetrator is not yet known). As a result, the undirected version of \textrank tends to favor the first part of the story and the resulting summary consists mainly of initial scenes. By adding extra importance to later scenes, we also encourage the selection of later events that might be surprising (and hence have lower similarity with other scenes) but more informative for the summary. 
Moreover, in \summer, where the weights change in a systematic manner based on narrative structure, we also observe that scenes appearing later in the screenplay are selected more often for inclusion in the summary.

As described in detail in Section 3.3, we also infer the narrative
structure of CSI episodes in the supervised version of \summer via
latent TP representations. During experimentation (see Section~5), we
found that these TPs are highly correlated with different aspects of a
CSI summary. In Figure~\ref{fig:TP_examples} we visualize examples of
identified TPs on CSI episodes during test time alongside with
gold-standard aspect-based summary annotations. Based on the examples,
we empirically observe that different TPs tend to capture different
types of information helpful for summarizing crime investigation
stories (e.g.,~crime scene, victim, perpetrator, motive).

\end{document}